\documentclass[sigconf]{acmart}
\renewcommand\footnotetextcopyrightpermission[1]{} 
\usepackage{multirow}
\usepackage{graphics}
\usepackage{graphicx}
\usepackage{subcaption}
\usepackage{tabularx}
\usepackage{color}
\usepackage{amsmath}
\usepackage{hyperref}
\usepackage{makecell}
\usepackage{caption}
\usepackage[ruled,vlined,linesnumbered]{algorithm2e}
\usepackage[T1]{fontenc}
\usepackage[english]{babel}
\begin{document}

\title{\emph{RheFrameDetect}: A Text Classification System for Automatic Detection of Rhetorical Frames in AI from Open Sources}

\author{Saurav Ghosh}
\affiliation{%
  \institution{Vista Consulting LLC}
  \city{Arlington}
  \state{Virginia}
  \country{USA}
}
\email{saurav.ghosh.vt@gmail.com}

\author{Philippe Loustaunau}
\affiliation{%
  \institution{Vista Consulting LLC}
  \city{Arlington}
  \state{Virginia}
  \country{USA}}
\email{ploustaunau@conseil-vista.com}


\begin{abstract}
  \emph{Rhetorical Frames} in AI can be thought of as expressions that describe AI development as a competition between two or more actors, such as governments or companies. Examples of such \emph{Frames} include \emph{robotic arms race}, \emph{AI rivalry}, \emph{technological supremacy}, \emph{cyberwarfare dominance} and \emph{5G race}. Detection of \emph{Rhetorical Frames} from open sources can help us track the attitudes of governments or companies towards AI, specifically whether attitudes are becoming more cooperative or competitive over time. Given the rapidly increasing volumes of open sources (online news media, twitter, blogs), it is difficult for subject matter experts to identify \emph{Rhetorical Frames} in (near) real-time. Moreover, these sources are in general unstructured (noisy) and therefore, detecting \emph{Frames} from these sources will require state-of-the-art text classification techniques. In this paper, we develop \emph{RheFrameDetect}, a text classification system for (near) real-time capture of \emph{Rhetorical Frames} from open sources. Given an input document, \emph{RheFrameDetect} employs text classification techniques at multiple levels (document level and paragraph level) to identify all occurrences of \emph{Frames} used in the discussion of AI. We performed extensive evaluation of the text classification techniques used in \emph{RheFrameDetect} against human annotated \emph{Frames} from multiple news sources. To further demonstrate the effectiveness of \emph{RheFrameDetect},  we show multiple case studies depicting the \emph{Frames} identified by \emph{RheFrameDetect} compared against human annotated \emph{Frames}.
\end{abstract}



\keywords{\emph{Rhetorical Frames}, Document Classification, Paragraph Classification, Doc2Vec, Paragraph Vector, Self-Attention, Attention Networks, \emph{FrameSpan} Identification}


\maketitle

\section{Introduction}
Advances in Artificial Intelligence (AI) have at times been described as an \emph{AI development race} among major nations in the world. World leaders, such as Russian President \emph{Vladimir Putin}, Chinese President \emph{Xi Jinping} and Chief Technology Officer of The Trump Administration have issued statements in the past calling for greater prioritization, swift action and increased investment in order to gain strategic advantage in this \emph{AI development race}. This paper focuses on how AI development is framed as a competition, either between nations or private companies. Our goal is to capture the \emph{Rhetorical Frames} of \emph{Competition} between governments or companies in the field of Artificial Intelligence (AI). This \emph{Frame} may be expressed in military terms (such as, \emph{arms race}, \emph{Cold War}, \emph{battle for supremacy}, etc.) or non-military (such as, \emph{win}, \emph{versus}, \emph{race}, \emph{compete with}, etc.). In Table~\ref{tab:frame_examples}, we show some examples and counter-examples of \emph{Rhetorical Frames} as well as the rationale behind them. For a more detailed understanding of \emph{Rhetorical Frames}, please refer to Imbrie et al.~\cite{RhetoricalCSET}

With massive online availability of open sources (online news media, twitter, blogs), identifying \emph{Rhetorical Frames} using human annotators can be very limited in scope as well as difficult to scale in real-time. Detecting \emph{Rhetorical Frames} from open sources in (near) real-time can assist policy makers or analysts in assessing whether attitudes of world governments or private companies towards AI development are becoming more cooperative or competitive over time. However, these textual open sources are unstructured (noisy) and therefore, pose challenges in identifying the occurrence of \emph{Frames}.

The availability of massive textual open sources coincides with recent developments in language modeling, such as Word2Vec~\cite{mikolovefficient,mikolovdistributed}, GloVe~\cite{Glovevector}, Doc2Vec~\cite{le2014distributed}, Sentence-BERT (\emph{SBERT})~\cite{reimers2019sentence} and others. These neural network based language models when trained over a representative corpus can be used to convert words as well as any piece of text (documents, paragraphs and sentences) to dense low-dimensional vector representations, most popularly known as embeddings. These embeddings are then provided as input to downstream classifiers for classifying a piece of text.

Motivated by these techniques, we develop \emph{RheFrameDetect}, a text classification system for automatic identification of \emph{Rhetorical Frames} from textual open sources. Our main contributions are as follows:
\vspace{-0.5em}
\begin{itemize}
    \item \textbf{Automated}: \emph{RheFrameDetect} is fully automatic. Given an input news article, it will automatically locate all the occurrences of \emph{Frames} mentioned in the context of AI.
    \item \textbf{Novelty}: To the best of our knowledge, there has been no prior systematic efforts at automatic detection of \emph{Rhetorical Frames} from open sources. 
    \item \textbf{Real-time}: \emph{RheFrameDetect} is an end-to-end system and can be deployed in a (near) real-time setting.
    \item \textbf{Evaluation}: We present a detailed and prospective analysis of \emph{RheFrameDetect} by evaluating the accuracies of text classifiers employed at document as well as paragraph level against manually annotated \emph{Frames} at the corresponding level.
    \item \textbf{Case Studies}: We also present multiple case studies providing insights into how \emph{Frames} located by \emph{RheFrameDetect} in news articles compare against the human annotated ones. 
\end{itemize}

\begin{table*}[!ht]
\centering
\caption{Examples and Counter-Examples of \emph{Rhetorical Frames} mentioned in the context of AI and the rationale behind them}
\resizebox{\textwidth}{!}{%
\begin{tabular}{|c|c|c|}
\hline
Text & \emph{Frame}? & Rationale \\ \hline
\begin{tabular}[c]{@{}c@{}}It is a great power \emph{rivalry} focusing on \emph{technological supremacy} \\ and which side has the best development model\end{tabular} & Yes & \begin{tabular}[c]{@{}c@{}}The phrase \emph{rivalry focusing on technological supremacy} occurs in \\ a discussion about AI and indicates a competition between two or more actors\end{tabular} \\ \hline
\begin{tabular}[c]{@{}c@{}}Chinese AI company \emph{iFlyTek} often \emph{beats Facebook, Alphabet's DeepMind, and IBM's Watson} \\ in \emph{competitions} to process natural speech\end{tabular} & Yes & \begin{tabular}[c]{@{}c@{}}The phrase \emph{beats Facebook, Alphabet's DeepMind, and IBM'sWatson in competitions} \\ describes a competition between two or more actors in the context of AI.\end{tabular} \\ \hline
China is \emph{outpacing} other countries in the development of 5G today. & Yes & \begin{tabular}[c]{@{}c@{}}Although one country is \emph{outpacing}, this term entails a race, or competition \\ between more than one actor, with the relative gains accruing to one side over another.\end{tabular} \\ \hline
US-China \emph{trade war} is making china stronger & No & \begin{tabular}[c]{@{}c@{}}The phrase \emph{trade war} is NOT an AI competition frame, because\\ it is not used in the context of AI.\end{tabular} \\ \hline
\begin{tabular}[c]{@{}c@{}}His company finds itself \emph{under fire}, \emph{besieged} by a U.S. effort to get key allies \\ to ban its networking equipment.\end{tabular} & No & \begin{tabular}[c]{@{}c@{}}This is NOT an AI competition frame, because \emph{under fire} and \emph{besieged} only describe\\ the actions of one actor; they do not entail competition or relative advantage\end{tabular} \\ \hline
\end{tabular}%
}
\label{tab:frame_examples}
\end{table*}

\section{Problem Overview}

In this paper, we intend to focus on detecting as well as locating \emph{Rhetorical Frames} mentioned in the context of AI within an incoming news article. \emph{RheFrameDetect} aims to solve both these problems by performing multiple tasks in a hierarchical fashion as given below.

\begin{itemize}
    \item \emph{DocContainsAI}: If the incoming article contains a discussion of AI, then \emph{RheFrameDetect} assigns a value of \emph{Yes} to \emph{DocContainsAI}. Otherwise, \emph{RheFrameDetect} assigns the value of \emph{No} and do not proceed for the document.
    \item \emph{DocContainsFrame}: If the incoming article contains at least one occurrence of \emph{Frames}, then \emph{RheFrameDetect} assigns a value of \emph{Yes} to \emph{DocContainsFrame}. Otherwise, \emph{RheFrameDetect} assigns the value of \emph{No} and do not proceed further.
    \item \emph{ParContainsFrame}: If the incoming document contains both a discussion of AI and at least one occurrence of \emph{Frames}, \emph{RheFrameDetect} aims to identify the paragraphs of the article which contain the \emph{Frames}. Therefore, for each paragraph, \emph{RheFrameDetect} assigns a value of \emph{Yes} to \emph{ParContainsFrame} if the paragraph contains at least one occurrence of \emph{Frames}. Otherwise, \emph{RheFrameDetect} assigns \emph{No} to the paragraph.
    \item \emph{FrameSpan}: For the paragraphs assigned \emph{Yes} for \emph{ParContainsFrame}, \emph{RheFrameDetect} identifies all occurrences of the \emph{Frames} mentioned in the paragraph.
\end{itemize}

\begin{figure}[t!]
\centering
\includegraphics[width=\linewidth]{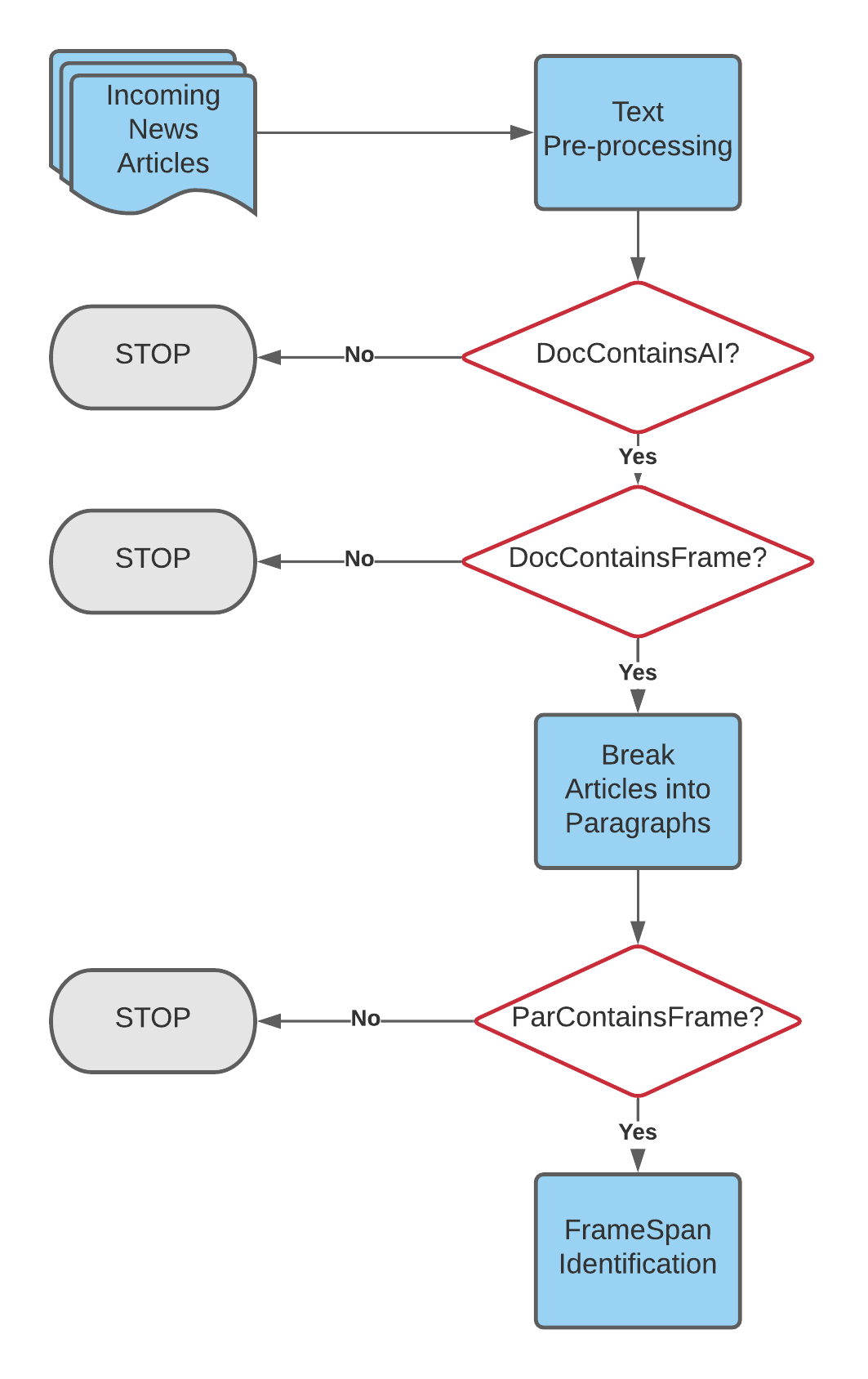}
\caption{Flow Chart depicting the tasks performed by \emph{RheFrameDetect} for incoming news articles as input up to \emph{FrameSpan} identification at the paragraph level.}
\label{fig:rheframeflowchart}
\end{figure}

A flowchart depicting the tasks performed by \emph{RheFrameDetect} for an incoming article is shown in Figure~\ref{fig:rheframeflowchart}.

\section{\emph{RheFrameDetect}}

In this section, we provide a brief description of the techniques used by \emph{RheFrameDetect} for solving the tasks shown in Figure~\ref{fig:rheframeflowchart}.

\subsection{\emph{DocContainsAI}}

Given an incoming news article, the first task that \emph{RheFrameDetect} performs is to detect whether the incoming article contains a discussion of AI. It is not necessary that AI be the main focus of the document. If there is any mention of AI or related keywords, then the field \emph{DocContainsAI} for the article is set to \emph{Yes}. For this task, \emph{RheFrameDetect} employs a standard keyword search technique to detect whether the article text contains the term \emph{AI} or related keywords.  

\subsection{\emph{DocContainsFrame}}

If \emph{DocContainsAI} is assigned the value of \emph{Yes}, then the next task performed by \emph{RheFrameDetect} is \emph{DocContainsFrame}. If the article contains at least one occurrence of the \emph{Rhetorical Frame} with reference to AI, the value \emph{Yes} is assigned to \emph{DocContainsFrame}, otherwise it is allocated the value \emph{No}. Therefore, \emph{DocContainsFrame} is a binary document classification task. To solve this task, \emph{RheFrameDetect} employs multiple classifiers as follows: \emph{Logistic Regression}, \emph{SVM}, \emph{Multi-layer Perceptron} and \emph{Random Forest}. However, all these classifiers require converting the input document into a low-dimensional feature embedding. For generating document embeddings, \emph{RheFrameDetect} employs two most popular and efficient techniques, namely \emph{Doc2Vec}~\cite{le2014distributed} and \emph{SBERT}~\cite{reimers2019sentence} for converting documents into real-valued embeddings in a low dimensional space. Document embeddings generated by \emph{Doc2Vec} and \emph{SBERT} are provided as input features to the downstream Machine Learning classifiers (\emph{Logistic Regression}, \emph{SVM}, \emph{Random Forest} and \emph{Multi-layer Perceptron}) for classifying \emph{DocContainsFrame}.

\subsection{\emph{ParContainsFrame}}

If \emph{DocContainsFrame} is assigned the value of \emph{Yes},  then the final binary classification task executed by \emph{RheFrameDetect} is \emph{ParContainsFrame} after segmenting the article into paragraphs. If each paragraph contains an instance of the \emph{Frame}, \emph{RheFrameDetect} assigns \emph{Yes} to \emph{ParContainsFrame}, otherwise \emph{No} is assigned to \emph{ParContainsFrame}. For \emph{ParContainsFrame}, \emph{RheFrameDetect} employs two types of classifiers as given below.

\begin{itemize}
    \item \textbf{Paragraph Vector based Classifiers:} One of the properties of the \emph{Doc2Vec} algorithm that make it so versatile, unique and powerful is that it can convert text (entire document or paragraph or sentence) of any length to low-dimensional real-valued embeddings. The same goes for the \emph{SBERT} algorithm in that it can convert documents or paragraphs or sentences of any length into low-dimensional vectors. Therefore, both \emph{Doc2Vec} and \emph{SBERT} are applicable to paragraphs too. Paragraph vectors generated by \emph{Doc2Vec} and \emph{SBERT} algorithms are provided as input features to the downstream classifiers (\emph{Logistic Regression}, \emph{SVM}, \emph{Random Forest} and \emph{Multi-layer Perceptron}) for classifying \emph{ParContainsFrame}.
    \item \textbf{Self-Attention based Classifiers:}  Self-Attention~\cite{lin2017structured} is an intra-attention mechanism that allows the words in a sentence or paragraph to interact with each other and discover which words in the sentence or paragraph should be paid more attention. \emph{RheFrameDetect} applies Self-Attention mechanism on top of bi-directional \emph{LSTM}~\cite{LSTMcite} for classifying \emph{ParContainsFrame}. Apart from classification, Self-Attention mechanism also allows us to locate the most important words in paragraphs in terms of contribution to the classification decision. Self-Attention based Classifiers take as input pre-trained GloVe~\cite{Glovevector} embeddings.
\end{itemize}

\subsection{\emph{FrameSpan} Identification (\emph{GuidedSelfAttention})}

If \emph{ParContainsFrame} is assigned a value of \emph{Yes}, \emph{RheFrameDetect} attempts to identify the \emph{FrameSpan} within the paragraph as the final task in the overall process of detection of \emph{Frames} from incoming news articles. Motivated by Rei et al.~\cite{rei2019jointly}, \emph{RheFrameDetect} employs a modified version of Self-Attention mechanism, namely \emph{GuidedSelfAttention} on top of bi-directional \emph{LSTM} to accomplish the task of identifying \emph{FrameSpan}. In \emph{GuidedSelfAttention}, the attention mechanism is modified in order to guide the attention values based on existing word-level \emph{FrameSpan} annotations as follows. 

\begin{itemize}
    \item Firstly, \emph{GuidedSelfAttention} converts the word-level \emph{FrameSpan} annotation within a paragraph to a 0-1 encoding, E.g., a paragraph of five words with first and third word being \emph{Frames} will be represented as $[1, 0, 1, 0, 0]$.
    \item In the second step, the 0-1 encoding corresponding to word-level \emph{FrameSpan} annotation is transformed into a probability distribution via normalization. E.g., $[1, 0, 1, 0, 0]$ is converted to $[0.5, 0, 0.5, 0, 0]$.
    \item Finally, \emph{GuidedSelfAttention} minimizes the KL divergence between the probability distributions of attention weights and word-level \emph{FrameSpan} annotation. As KL divergence is a measure of the difference between two probability distributions, minimizing it will encourage \emph{GuidedSelfAttention} to apply more attention to the areas in paragraphs containing the \emph{FrameSpan}.
\end{itemize}

Therefore, \emph{GuidedSelfAttention} optimizes an additional objective of KL divergence between attention weights distribution and \emph{FrameSpan} annotation distribution, apart from the \emph{ParContainsFrame} classification objective. Minimizing this joint objective encourages \emph{GuidedSelfAttention} to focus more on the tokens within \emph{FrameSpan} by applying more attention weights to them.  

\section{Experimental Evaluation}

In this section, we provide a brief description of our experimental setup, including the dataset of news articles, the annotation process and the classifiers used by \emph{RheFrameDetect} for the tasks \emph{DocContainsFrame} and \emph{ParContainsFrame}.

\subsection{News Articles Corpus}

To explore the use of \emph{Rhetorical Frames}, we analyzed a corpus of news articles related to AI from four different sources, namely \emph{Reuters News Agency}, \emph{Defense One}, \emph{Foreign Affairs} and \emph{LexisNexis news database}~\cite{LexisNexisDatabase} over the period 2012 to 2019. These news sources were selected based on three criteria: firstly, they have a broad, global news coverage; secondly, they offered a diversity of news analysis, commentary, and opinion; and thirdly, they are more consistently editorially structured than speeches, blogs, or social media posts. In Table~\ref{tab:annotation_counts}, we show the total number of news articles containing AI and \emph{Frames} as well as the total number of paragraphs across articles containing \emph{Frames} for each source. As can be seen in Table~\ref{tab:annotation_counts}, both binary classification tasks \emph{DocContainsFrame} and \emph{ParContainsFrame} have a class imbalance of 14:1 and 13:1 respectively, with \emph{Yes} being the minority class. Therefore, it is extremely challenging to predict whether a document or a paragraph contains \emph{Frames} as classification data is heavily skewed towards documents or paragraphs not having any \emph{Frames}. All articles were provided by the Center for Security and Emerging Technology (\href{https://cset.georgetown.edu/}{\emph{CSET}}) at Georgetown University~\cite{RhetoricalCSET}.

\subsection{Human Annotation}

Analysts at \emph{CSET} helped identify Rhetorical Frames in an initial small set of news articles. From this, we developed an initial annotation guide that we tested with our annotation service provider (\href{https://www.spi-global.com/}{\emph{SPi Global}}). This involved multiple back and forth with the analysts at \emph{CSET}. We finalized the annotation guidelines and provided ~19K documents to \emph{SPi Global} for annotation. We also asked \emph{SPi Global} to doubly and independently annotate 10\% of these documents to assess inter-coder agreement. For each article the annotators were asked to determine whether the article discussed AI, if so, whether it contained the \emph{Rhetorical Frame}. For documents that contained the \emph{Frame}, and for each paragraph in that document, the annotators were asked to determine whether the paragraph contained the \emph{Frame}, and identify the text span of all instances of the \emph{Frame}. The inter-coder agreement at the document level was 97\% and at the paragraph level was 87\%.

\begin{table*}[!ht]
\centering
\caption{Counts of articles for each source with \emph{DocContainsAI} assigned Yes or No and \emph{DocContainsFrame} assigned Yes or No. We also show the counts of paragraphs across articles for each source with \emph{ParContainsFrame} assigned Yes or No}
\resizebox{\textwidth}{!}{%
\begin{tabular}{|c|c|c|c|c|c|}
\hline
\emph{Classification Task} & \emph{Class} & \emph{Reuters} & \begin{tabular}[c]{@{}c@{}}\emph{Defense} \\ \emph{One}\end{tabular} & \begin{tabular}[c]{@{}c@{}}\emph{Foreign} \\ \emph{Affairs}\end{tabular} & \emph{LexisNexis} \\ \hline
\multirow{2}{*}{\emph{DocContainsAI}} & \emph{Yes} & 3496 & 537 & 55 & 9984 \\ \cline{2-6} 
 & \emph{No} & 4205 & 667 & 11 & 16 \\ \hline
\multirow{2}{*}{\emph{DocContainsFrame}} & \emph{Yes} & 249 & 43 & 1 & 649 \\ \cline{2-6} 
 & \emph{No} & 3247 & 494 & 54 & 9335 \\ \hline
\multirow{2}{*}{\emph{ParContainsFrame}} & \emph{Yes} & 391 & 79 & 1 & 1032 \\ \cline{2-6} 
 & \emph{No} & 4934 & 798 & 29 & 13998 \\ \hline
\end{tabular}%
}
\label{tab:annotation_counts}
\end{table*}

\subsection{Text Pre-processing}

The textual content of each article in the corpus was pre-processed using SpaCy~\cite{spacycite}, Gensim~\cite{rehurek_lrec}, NLTK~\cite{NLTKcite} and Keras~\cite{chollet2015keras}. Pre-processing steps involve sentence splitting, tokenization, token lower-casing, noise removal, and ignoring too short or too long tokens.

\subsection{Classifiers}

As per Figure~\ref{fig:rheframeflowchart}, we have two binary classification tasks: 1) \emph{DocContainsFrame} at the article level and 2) \emph{ParContainsFrame} at the paragraph level. 

For \emph{DocContainsFrame}, \emph{RheFrameDetect} evaluated two types of classifiers as follows.
\begin{itemize}
    \item \textbf{\emph{Doc2Vec} based Classifiers:} \emph{Doc2Vec} embeddings provided as input features to downstream classifiers, namely \emph{Logistic Regression}, \emph{SVM}, \emph{Random Forest} and \emph{Multi-layer Perceptron}. We compared four variants of Doc2Vec: 1) Distributed Bag of Words with hierarchical softmax (\emph{DV-DBOW-HS}) and negative sampling (\emph{DV-DBOW-NEG}), 2) Distributed Memory with hierarchical softmax (\emph{DV-DM-HS}) and Negative Sampling (\emph{DV-DM-NEG}). For each of these models, we used 300 embedding dimension, a window size of 5 and 5 negative samples if negative sampling is used. Our corpus of 18972 AI related news articles was used for training the Doc2Vec models to generate the embeddings. 
    \item \textbf{\emph{SBERT} based Classifiers:} \emph{SBERT} embeddings for documents provided as input features to downstream classifiers, namely \emph{Logistic Regression}, \emph{SVM}, \emph{Random Forest} and \emph{Multi-layer Perceptron}. For \emph{SBERT} model, we used 768 embedding dimension. We used the implementation of \emph{SBERT} model \emph{all-mpnet-base-v2} within the \emph{SentenceTransformer} package of \emph{HuggingFace}. \emph{SBERT} model \emph{all-mpnet-base-v2} maps each sentence/paragraph/document to a dense 768-dimensional vector space and was trained on a large and diverse dataset of over 1 billion training sentence pairs. 
    
\end{itemize}

For \emph{ParContainsFrame}, \emph{RheFrameDetect} compared the following classifiers.

\begin{itemize}
    \item \textbf{\emph{Paragraph Vector} based Classifiers:} \emph{Paragraph Vectors}~\cite{le2014distributed} provided as input features to the same downstream classifiers as used in \emph{Doc2Vec} based classifiers for \emph{DocContainsFrame}. Similar to \emph{Doc2Vec}, we also compared four variants of \emph{Paragraph Vector}: \emph{PV-DBOW-HS}, \emph{PV-DBOW-NEG}, \emph{PV-DM-HS} and \emph{PV-DM-NEG}. We used the same parameter settings for generation of \emph{Paragraph Vectors} (300 embedding dimension, window size of 5 and 5 negative samples if negative sampling is used) as in \emph{Doc2Vec}. We extracted all the sentences from our corpus of 18972 AI related news articles and generated a sentence corpus comprising of nearly 1 million sentences for training the \emph{Paragraph Vector} models.
    \item \textbf{\emph{SBERT} based classifiers:} \emph{SBERT} 768-dimensional embeddings for paragraphs generated using \emph{all-mpnet-base-v2} model within the \emph{SentenceTransformer} package of \emph{HuggingFace} provided as input features to the same downstream classifiers as used for \emph{DocContainsFrame}. 
    \item \textbf{\emph{SelfAttention}:} Self-Attention mechanism~\cite{lin2017structured} that learns \emph{ParContainsFrame} with \emph{GloVe} word embeddings as input.
    \item \textbf{\emph{GuidedSelfAttention}:} Self-Attention mechanism~\cite{lin2017structured} that jointly learns \emph{ParContainsFrame} and \emph{FrameSpan} with \emph{GloVe} word embeddings as input.
\end{itemize}

\emph{Doc2Vec} and \emph{Paragraph Vectors} were implemented using \emph{Gensim}~\cite{rehurek_lrec}. \emph{SBERT} embeddings were generated using the \emph{SentenceTransformer} package of \emph{HuggingFace}.  For each of the four downstream classifiers (\emph{Logistic Regression}, \emph{SVM}, \emph{Random Forest} and \emph{Multi-layer Perceptron}), we used a grid of parameter settings as shown below and then used \emph{scikit-learn}~\cite{scikit-learn}'s \emph{GridSearchCV} (10-fold cross-validation) to perform an exhaustive search over the parameter grid for each classifier.
\begin{itemize}
    \item \textbf{\emph{Logistic Regression}:} \emph{penalty}: [`l1', `l2', `elasticnet'], \emph{C}:[0.1, 1., 10., 100., 1000.]
    \item \textbf{\emph{SVM}:} \emph{kernel}: [`rbf', `poly', `linear'], \emph{C}:[0.1, 1., 10., 100., 1000.], \emph{degree}: [3, 5, 7, 8]
    \item \textbf{\emph{Multi-layer Perceptron}:} \emph{hiddenlayersizes}: [(100,)], \emph{learningrate}: [`constant', `invscaling', `adaptive'], \emph{activation}: [`logistic', `identity', `tanh', `relu'], \emph{alpha}: [0.1, 0.01, 0.001, 0.0001]
    \item \textbf{\emph{Random Forest}:} \emph{nestimators}: [10, 50, 100, 200, 500, 1000], \emph{maxfeatures}: [`log2', `sqrt']
\end{itemize}

For all the other parameters (not part of \emph{GridSearchCV}) in each classifier, we used the default parameter setting as in \emph{scikit-learn}. The \emph{GridSearchCV} procedure gives us the optimal parameter setting for each downstream classifier and we refit each classifier on the entire training set using the optimal parameter setting following the \emph{GridSearchCV} procedure. To account for the class imbalance in both \emph{DocContainsFrame} and \emph{ParContainsFrame}, we used the parameter \emph{classweight}: \emph{balanced} in all of the classifiers. \emph{SelfAttention} and \emph{GuidedSelfAttention}) were implemented using Keras~\cite{chollet2015keras} and Tensorflow~\cite{tensorflowcite}. For both \emph{SelfAttention} and \emph{GuidedSelfAttention}, we employed \emph{EarlyStopping} for monitoring \emph{crossentropy} loss on the validation dataset (10 percent of training set). If the validation loss is no longer decreasing for 2-3 iterations consecutively, we stop the training, thereby minimizing the risk of potential overfitting for both \emph{SelfAttention} and \emph{GuidedSelfAttention}.

\subsection{Performance Metrics}

We evaluated the classifiers for each binary classification task (\emph{DocContainsFrame} and \emph{ParContainsFrame}) in terms of the following performance metrics.
\begin{itemize}
    \item \textbf{\emph{Precision}} for each class (\emph{Yes} and \emph{No}) as well as macro-average across the classes
    \item \textbf{\emph{Recall}} for each class as well as macro-average across the classes
    \item \textbf{\emph{F1-score}} for each class as well as macro-average across the classes
\end{itemize}

We used \emph{scikit-learn}'s Repeated Stratified 5-Fold cross validator for splitting the annotation data corresponding to each binary classification task into 5 folds. For each fold, we calculate the score for each metric and report the average metric score across the folds.  

\section{Results}

In this section, we try to ascertain the efficacy and applicability of \emph{RheFrameDetect} by investigating some of the pertinent questions related to the problem of automated detection of \emph{Rhetorical Frames}.

\par{\textbf{\emph{Doc2Vec} based Classifiers vs \emph{SBERT} based classifiers - Which is a better method for detecting \emph{Rhetorical Frames} at the document level?}}

In Tables~\ref{tab:doccontainsframe1} and~\ref{tab:doccontainsframe2}, we evaluated \emph{Doc2Vec} and \emph{SBERT} based classifiers for the binary classification task \emph{DocContainsFrame}. For \emph{DocContainsFrame}, we are mostly interested in the performance metrics (\emph{Precision}, \emph{Recall} and \emph{F1-score}) for \emph{Yes} class as it is the minority class and therefore, hard to predict. Moreover, performance metrics for the \emph{Yes} class will inform us whether classifiers can detect presence of \emph{Frames} at the document level. We are also interested in the macro-average metrics across the classes. Overall, in terms of \emph{F1-score} for \emph{Yes} class and macro-average, \emph{Multi-layer Perceptron} with embedding \emph{DV-DBOW-NEG} and \emph{SVM} with embedding \emph{SBERT} emerge out to be the best performing classifiers. In terms of \emph{Recall}, \emph{Logistic Regression} consistently emerges out to be the best classifier but performs worse in terms of \emph{Precision} leading to low \emph{F1-score} than \emph{SVM} and \emph{Multi-layer Perceptron}. \emph{Random Forest} outperforms other classifiers in terms of \emph{Precision} but performs worse in terms of \emph{Recall} leading to very poor \emph{F1-score} compared to \emph{SVM} and \emph{Multi-layer Perceptron}.  

If we compare \emph{Doc2Vec} vs \emph{SBERT} in terms of the \emph{F1-score}, we do not notice any significant performance difference. However, in terms of \emph{Recall}, \emph{Logistic Regression} performs better with \emph{Doc2Vec} embeddings than \emph{SBERT}. Similarly, for \emph{Precision}, \emph{Random Forest} performs better with \emph{Doc2Vec} embeddings than \emph{SBERT}. This depicts that \emph{Doc2Vec} generates slightly better embeddings at the document level than \emph{SBERT}. Since \emph{SBERT} technique is trained over a data set of nearly billion sentences, it may not be the ideal technique for generating embeddings at the document level. 

\par{\textbf{\emph{Paragraph Vector} and \emph{SBERT} based Classifiers vs Self-Attention based Classifiers - Which is a better method for detecting \emph{Rhetorical Frames} at the paragraph level?}}

In Tables~\ref{tab:parcontainsframe1}, ~\ref{tab:parcontainsframe2} and ~\ref{tab:parcontainsframedeep}, we evaluated \emph{Paragraph Vector} and \emph{SBERT} based classifiers against the Self-Attention based classifiers for the binary classification task \emph{ParContainsFrame}. 

Firstly, we compare \emph{Paragraph Vector} based classifiers against \emph{SBERT} based classifiers for this task. Similar to \emph{DocContainsFrame}, we are mostly interested in the performance metrics (\emph{Precision}, \emph{Recall} and \emph{F1-score}) for \emph{Yes} class as it is the minority class and therefore, hard to predict. Moreover, performance metrics for the \emph{Yes} class will inform us whether classifiers can detect presence of \emph{Frames} at the paragraph level. As can be clearly seen, \emph{SVM} with \emph{SBERT} embeddings outperform all the other techniques in terms of \emph{F1-Score} for the \emph{Yes} class as well as macro-average. Overall, with both \emph{Doc2Vec} and \emph{SBERT} embeddings, \emph{Multi-layer Perceptron} emerges out to be the best performing classifier in terms of \emph{F1-Score} followed by \emph{SVM}. Similar to \emph{DocContainsFrame}, \emph{Logistic Regression} is the best performing classifier in terms of \emph{Recall} for \emph{ParContainsFrame} and \emph{Random Forest} outperforms other classifiers in terms of \emph{Precision}.

When we compared \emph{Paragraph Vector} against \emph{SBERT} across the four classifiers, \emph{SBERT} based classifiers provide better \emph{F1-score} than the corresponding \emph{Paragraph Vector} based classifiers. \emph{SBERT} model is trained over a dataset of billion sentences in comparison to \emph{Paragraph Vector} trained over our corpus dataset of nearly 1 million sentences and therefore, \emph{SBERT} model is expected to generate better quality embeddings for sentences/paragraphs than \emph{Paragraph Vector}.

Finally, we compared the Self-Attention based classifiers (\emph{SelfAttention} and \emph{GuidedSelfAttention}) against the \emph{Paragraph Vector} and \emph{SBERT} based classifiers. As expected, Self-Attention based classifiers (specifically \emph{GuidedSelfAttention}) outperform \emph{Paragraph Vector} and \emph{SBERT} based classifiers in terms of \emph{Recall} and \emph{F1-Score} due to the powerful Self-Attention mechanism guided by the \emph{FrameSpan} annotations. For the \emph{Yes} class of \emph{ParContainsFrame}, \emph{GuidedSelfAttention} achieves the highest \emph{Recall} (0.87) and \emph{F1-Score} (0.81) among all the classifiers. Superior performance of Self-Attention based classifiers is expected as they are sequence based models. Therefore, they learn the sequential patterns of Frames in paragraphs more efficiently compared to the Paragraph Vector based classifiers. Moreover, Self-Attention mechanism provides dual benefits to the classifier: not only does it result in better classification metrics, but it also drives the classifier to locate the most relevant words in paragraphs/sentences which contribute to the classification decision.

\section{Case Studies: \emph{FrameSpan} Identification}

\begin{figure*}[ht!]
  \centering
  \begin{subfigure}{0.45\textwidth}
    \centering
    \includegraphics[width=0.90\linewidth]{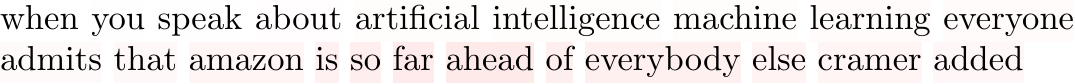}
    \caption{Case Study 1 (\emph{SelfAttention})}
  \end{subfigure}
  \begin{subfigure}{0.45\textwidth}
    \centering
    \includegraphics[width=0.90\linewidth]{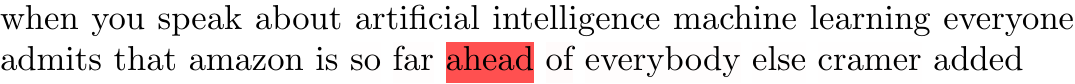}
    \caption{Case Study 1 (\emph{GuidedSelfAttention})}
  \end{subfigure}
  \\
  \begin{subfigure}{0.45\textwidth}
    \centering
    \includegraphics[width=0.90\linewidth]{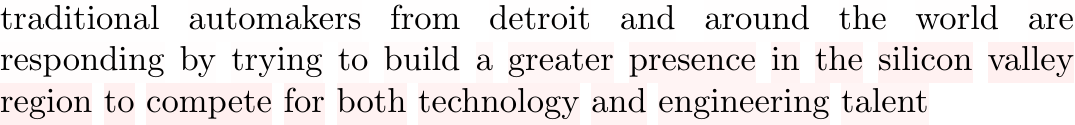}
    \caption{Case Study 2 (\emph{SelfAttention})}
    \label{fig:outcome}
  \end{subfigure}
  \begin{subfigure}{0.45\textwidth}
    \centering
    \includegraphics[width=0.90\linewidth]{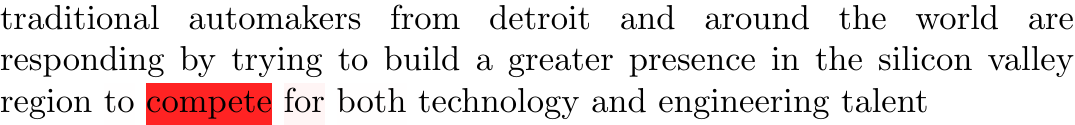}
    \caption{Case Study 2 (\emph{GuidedSelfAttention})}
    \label{fig:animal}
  \end{subfigure}
  \\
  \begin{subfigure}{0.45\textwidth}
    \centering
    \includegraphics[width=0.90\linewidth]{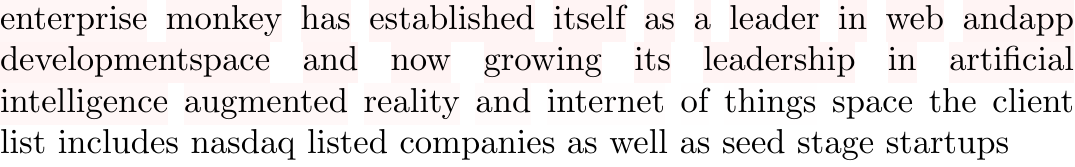}
    \caption{Case Study 3 (\emph{SelfAttention})}
    \label{fig:comorbidities}
  \end{subfigure}
  \begin{subfigure}{0.45\textwidth}
    \centering
    \includegraphics[width=0.90\linewidth]{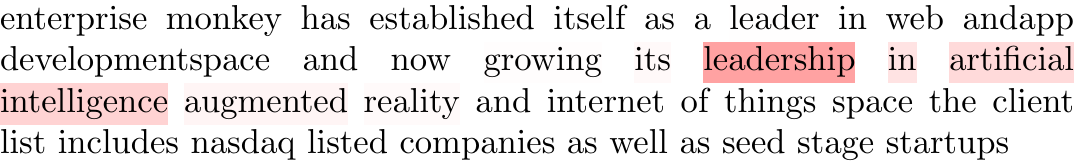}
    \caption{Case Study 3 (\emph{GuidedSelfAttention})}
    \label{fig:HCW}
  \end{subfigure}
  \\
  \begin{subfigure}{0.45\textwidth}
    \centering
    \includegraphics[width=0.90\linewidth]{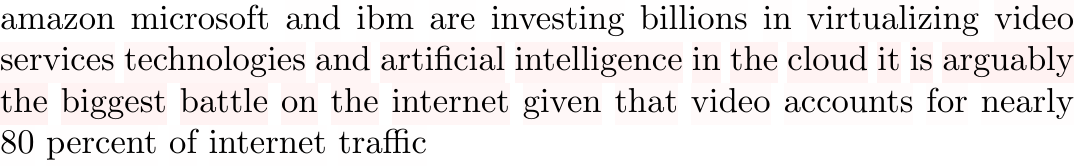}
    \caption{Case Study 4 (\emph{SelfAttention})}
    \label{fig:comorbidities}
  \end{subfigure}
  \begin{subfigure}{0.45\textwidth}
    \centering
    \includegraphics[width=0.90\linewidth]{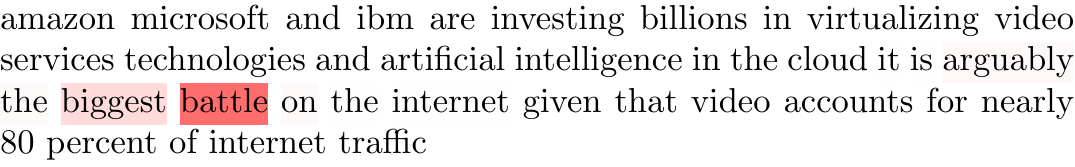}
    \caption{Case Study 4 (\emph{GuidedSelfAttention})}
  \end{subfigure}
  \\
  \begin{subfigure}{0.45\textwidth}
    \centering
    \includegraphics[width=0.90\linewidth]{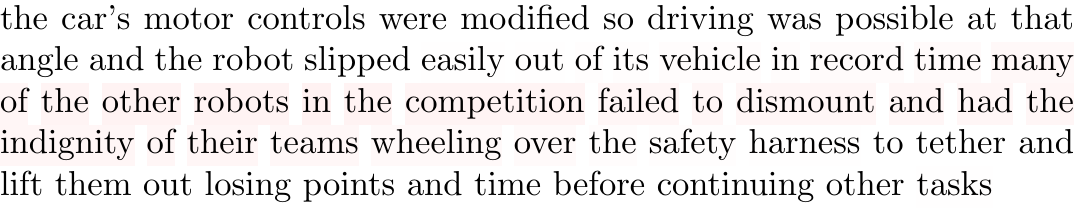}
    \caption{Case Study 5 (\emph{SelfAttention})}
  \end{subfigure}
  \begin{subfigure}{0.45\textwidth}
    \centering
    \includegraphics[width=0.90\linewidth]{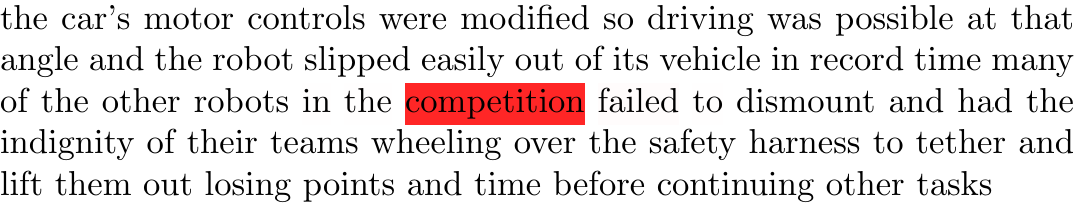}
    \caption{Case Study 5 (\emph{GuidedSelfAttention})}
  \end{subfigure}
  \caption{Comparing the weights of words in paragraphs for each case study as shown in Table~\ref{tab:caseStudyFrameSpan} assigned by \emph{SelfAttention} and \emph{GuidedSelfAttention} models. Words in paragraphs are highlighted corresponding to their weights. We can see that \emph{GuidedSelfAttention} is able to locate the \emph{Frames} in each paragraph more accurately than \emph{SelfAttention}.}
  \label{fig:caseStudyAttention}
\end{figure*}

Finally, we also depict five case studies related to \emph{FrameSpan} identification after \emph{ParContainsFrame} classification. For each case study, we chose a paragraph from the test dataset of \emph{ParContainsFrame} classification and then, we visualized the attention weights of tokens in each paragraph assigned by \emph{SelfAttention} and \emph{GuidedSelfAttention} models. In Table~\ref{tab:caseStudyFrameSpan}, we show the selected paragraphs and the \emph{FrameSpan} mentioned within each paragraph. In Figure~\ref{fig:caseStudyAttention}, we show the visualization of the attention weights for both \emph{SelfAttention} and \emph{GuidedSelfAttention}. As can be seen clearly in Figure~\ref{fig:caseStudyAttention}, \emph{GuidedSelfAttention} is able to locate the tokens within \emph{FrameSpan} more prominently in comparison to \emph{SelfAttention}. Specifically, in paragraphs corresponding to case studies 2 and 5, \emph{GuidedSelfAttention} captures accurately the \emph{FrameSpan} \emph{`compete'} and \emph{`competition'} respectively. This is expected as the attention weights in \emph{GuidedSelfAttention} are supervised based on the \emph{FrameSpan} annotations. This motivates \emph{GuidedSelfAttention} to apply more attention to the portions in paragraphs which may contain \emph{FrameSpan}. On the other hand, attention weights in \emph{SelfAttention} model are unsupervised, i.e. it does not have any information about the \emph{FrameSpan} location in paragraphs. Therefore, \emph{GuidedSelfAttention} not only outperforms \emph{SelfAttention} in terms of the metrics (\emph{Recall} and \emph{F1-Score}) for the \emph{Yes} class in \emph{ParContainsFrame} but also has the capability to accurately locate \emph{FrameSpan} information within the paragraph leading to enhanced intelligibility from a human analyst perspective.

\begin{table*}[!ht]
\scriptsize
\centering
\caption{Case studies for \emph{FrameSpan} identification where we show the paragraphs and the \emph{Frames} mentioned within the paragraphs for each case study}
\begin{tabular}{|c|c|c|}
\hline
Case Study & Paragraph                                                                                                                                                                                                                                                                                                                                                                                                                      & \emph{FrameSpan}                               \\ \hline
1          & \textit{\begin{tabular}[c]{@{}c@{}}"When you speak about artificial intelligence, machine learning, \\ everyone admits that...(Amazon) is so far ahead of everybody else," Cramer added.\end{tabular}}                                                                                                                                                                                                              & \textit{ahead of everybody else}               \\ \hline
2          & \textit{\begin{tabular}[c]{@{}c@{}}Traditional automakers from Detroit and around the world are responding \\ by trying to build a greater presence in the Silicon Valley region, \\ to compete for both technology and engineering talent.\end{tabular}}                                                                                                                                                                                                                                                                                                       & \textit{compete}                   \\ \hline
3          & \textit{\begin{tabular}[c]{@{}c@{}}Enterprise Monkey has established itself as a leader in web and app development space \\ and now growing its leadership in Artificial Intelligence, Augmented Reality and Internet of Things space. \\ The client list includes NASDAQ listed companies as well as seed-stage startups.\end{tabular}}                                                                                                                                                                                                                        &   \textit{growing its leadership in Artificial Intelligence}        \\ \hline
4          & \textit{\begin{tabular}[c]{@{}c@{}}Amazon, Microsoft and IBM are investing billions in virtualizing video services \\ (technologies) and Artificial Intelligence in the cloud. It is arguably the biggest battle on the internet, \\ given that video accounts for nearly 80 percent of internet traffic.\end{tabular}}                                                                                                                                                                                                                                                                                                                         & \textit{biggest battle on the internet}                        \\ \hline
5          & \textit{\begin{tabular}[c]{@{}c@{}}The car's motor controls were modified so driving was possible at that angle \\ and the robot slipped easily out of its vehicle in record time. \\ Many of the other robots in the competition failed to dismount \\ and had the indignity of their teams wheeling over the safety harness to tether and lift them out, \\ losing points and time, before continuing other tasks.\end{tabular}} & \textit{competition} \\ \hline
\end{tabular}
\label{tab:caseStudyFrameSpan}
\end{table*}

\vspace{-1em}
\section{Conclusions and Future Work}

In this manuscript, we have introduced \emph{RheFrameDetect}, the first automated end-to-end system for detecting and locating \emph{Rhetorical Frames} in AI from Open Sources. \emph{RheFrameDetect} takes an incoming news article as input and performs multiple tasks (Text Pre-processing, Keyword matching and binary classifications) in a hierarchical manner. Binary classification tasks performed by \emph{RheFrameDetect} were used to detect Frames at both document level (\emph{DocContainsFrame}) and paragraph level (\emph{ParContainsFrame}). We evaluated a wide variety of classifiers for both these tasks. For \emph{DocContainsFrame} classification task, we found that \emph{Multi-layer Perceptron} and \emph{SVM} emerge out to be the best performing classifiers in terms of \emph{F1-Score} for the \emph{Yes} class and macro-average. In terms of \emph{Recall} and \emph{Precision} for \emph{DocContainsFrame}, \emph{Doc2Vec} based classifiers perform slightly better than \emph{SBERT} based classifiers indicating better quality of embeddings generated at the document level by \emph{Doc2Vec} than \emph{SBERT}. For the \emph{ParContainsFrame} classification task, \emph{GuidedSelfAttention} classifier outperforms all other classifiers achieving a macro-average \emph{F1-Score} of 0.90. For the \emph{Yes} class, \emph{GuidedSelfAttention} achieves the highest \emph{Recall} of 0.87 and \emph{F1-Score} of 0.81. We showed in Figure~\ref{fig:caseStudyAttention} that \emph{GuidedSelfAttention} also has the capability to accurately locate \emph{FrameSpan} information within the paragraph compared to \emph{SelfAttention}. Superior performance of \emph{GuidedSelfAttention} can be attributed to the fact that it directs the Self-Attention mechanism to put more emphasis (higher weights) on words/tokens within \emph{FrameSpan} during training, ignoring the other words/tokens in paragraphs.

Our future work will focus on expanding \emph{RheFrameDetect} to detect the purpose of \emph{Rhetorical Frames} in AI, such as \emph{Motivation}, \emph{Critique}, \emph{Structured} and \emph{Explanation}. We also aim to quantify the \emph{FrameSpan} identification task in terms of \emph{Precision}, \emph{Recall} and \emph{F1-Score} using rankings based on attention weights of words/tokens in each paragraph. Finally, we also aim to extract named entities mentioned in the context of \emph{Frames} within a paragraph so that we can provide human analysts more context around the detected \emph{Frames}.

\begin{table*}[!ht]
\tiny
\centering
\caption{Comparing \emph{Doc2Vec} (\emph{DV-DM-NEG} and \emph{DV-DBOW-NEG}) based Classifiers for \emph{DocContainsFrame} in terms of \emph{Precision}, \emph{Recall} and \emph{F1-score} for each class (\emph{Yes} and \emph{No}) as well as macro-average across the classes.}
\resizebox{\textwidth}{!}{%
\begin{tabular}{|c|c|c|c|c|c|}
\hline
\emph{Embedding} & \emph{Classifier} & \emph{Class}         & \emph{Precision} & \emph{Recall} & \emph{F1-score} \\ \hline
\multirow{12}{*}{\emph{DV-DM-NEG}} & \multirow{3}{*}{\emph{Logistic Regression}} & \emph{No}            & 0.97      & 0.75   & 0.84     \\ 
                                   &                   & \emph{Yes}           & 0.15      & \textbf{0.63}   & 0.24     \\ 
                                   &                   & \emph{Macro-average} & 0.56      & \textbf{0.69}   & 0.54     \\ \cline{2-6}
                                  & \multirow{3}{*}{\emph{SVM}}               & \emph{No}            & 0.95      & 0.93   & 0.94     \\  
                                   &                   & \emph{Yes}           & 0.24      & 0.29   & \textbf{0.26}     \\  
                                  &                   & \emph{Macro-average} & 0.59      & 0.61   & \textbf{0.60}     \\ \cline{2-6}
                                & \multirow{3}{*}{\emph{Multi-layer Perceptron}}   & \emph{No}            & 0.94      & 0.99   & 0.97  \\ 
                                  &                    & \emph{Yes}           & 0.53      & 0.08   & 0.13     \\  
                                  &                   & \emph{Macro-average} & 0.73      & 0.54   & 0.55     \\ \cline{2-6}
                                & \multirow{3}{*}{\emph{Random Forest}}     & \emph{No}            & 0.93      & 1.0   & 0.97     \\  
                                  &                    & \emph{Yes}           & \textbf{1.0}      & 0.01   & 0.02     \\  
                                  &                    & \emph{Macro-average} & \textbf{0.97}      & 0.50   & 0.49     \\ \hline
\multirow{12}{*}{\emph{DV-DBOW-NEG}} & \multirow{3}{*}{\emph{Logistic Regression}}    & \emph{No}            & 0.97      & 0.77   & 0.86 \\  
                                   &                   & \emph{Yes}           & 0.17      & \textbf{0.67}   & 0.27     \\  
                                  &                      & \emph{Macro Average} & 0.57      & \textbf{0.72}   & 0.56     \\ \cline{2-6}
                                    & \multirow{3}{*}{\emph{SVM}}                   & \emph{No}      & 0.95      & 0.98   & 0.96     \\  
                                   &                   & \emph{Yes}           & 0.37      & 0.23   & 0.29     \\  
                                    &                  & \emph{Macro Average} & 0.66      & 0.61   & 0.63     \\ \cline{2-6}
                                & \multirow{3}{*}{\emph{Multi-layer Perceptron}} & \emph{No}    & 0.95      & 0.99   & 0.97     \\  
                                     &                 & \emph{Yes}           & 0.54      & 0.24   & \textbf{0.33}     \\  
                                      &                & \emph{Macro-average} & 0.75      & 0.61   & \textbf{0.65}     \\ \cline{2-6}
                                & \multirow{3}{*}{\emph{Random Forest}}   & \emph{No}            & 0.93      & 1.00   & 0.97     \\  
                                       &               & \emph{Yes}           & \textbf{0.85}      & 0.01   & 0.03     \\  
                                    &                  & \emph{Macro-average} & \textbf{0.89}      & 0.51   & 0.50     \\ \hline
\end{tabular}%
}
\label{tab:doccontainsframe1}
\end{table*}

\begin{table*}[!ht]
\tiny
\centering
\caption{Comparing \emph{Doc2Vec} (\emph{DV-DM-HS} and \emph{DV-DBOW-HS}) based Classifiers for \emph{DocContainsFrame} in terms of \emph{Precision}, \emph{Recall} and \emph{F1-score} for each class (\emph{Yes} and \emph{No}) as well as macro-average across the classes.}
\resizebox{\textwidth}{!}{%
\begin{tabular}{|c|c|c|c|c|c|}
\hline
\emph{Embedding} & \emph{Classifier} & \emph{Class}         & \emph{Precision} & \emph{Recall} & \emph{F1-score} \\ \hline
\multirow{12}{*}{\emph{DV-DM-HS}} & \multirow{3}{*}{\emph{Logistic Regression}} & \emph{No}          & 0.96      & 0.74   & 0.84     \\  
                                   &                   & \emph{Yes}           & 0.15      & \textbf{0.61}   & 0.24     \\  
                                    &                  & \emph{Macro-average} & 0.55      & \textbf{0.68}   & 0.54     \\ \cline{2-6}
                        & \multirow{3}{*}{\emph{SVM}}                       & \emph{No}            & 0.94      & 0.99   & 0.96     \\  
                                   &                   & \emph{Yes}           & 0.08      & 0.06   & 0.07     \\  
                                    &                  & \emph{Macro-average} & 0.51      & 0.53   & 0.52     \\ \cline{2-6}
                        & \multirow{3}{*}{\emph{Multi-layer Perceptron}}    & \emph{No}            & 0.95      & 0.99   & 0.97     \\  
                                   &                   & \emph{Yes}           & 0.51      & 0.20   & \textbf{0.29}     \\  
                                &                      & \emph{Macro-average} & 0.73      & 0.59   & \textbf{0.63}     \\ \cline{2-6}
                        & \multirow{3}{*}{\emph{Random Forest}}      & \emph{No}            & 0.93      & 1.00   & 0.97     \\  
                                  &                    & \emph{Yes}           & \textbf{0.97}      & 0.02   & 0.03     \\  
                                  &                    & \emph{Macro-average} & \textbf{0.95}      & 0.51   & 0.50     \\ \hline
\multirow{12}{*}{\emph{DV-DBOW-HS}} & \multirow{3}{*}{\emph{Logistic Regression}}     & \emph{No}            & 0.97      & 0.75   & 0.85 \\
                                   &                   & \emph{Yes}           & 0.16      & \textbf{0.66}   & 0.26     \\  
                                    &                  & \emph{Macro-average} & 0.57      & \textbf{0.71}   & 0.55     \\ \cline{2-6}
                        & \multirow{3}{*}{\emph{SVM}}                     & \emph{No}            & 0.95      & 0.98   & 0.96     \\  
                                    &                  & \emph{Yes}           & 0.37      & 0.23   & 0.28     \\  
                                    &                  & \emph{Macro-average} & 0.66      & 0.61   & 0.62     \\ \cline{2-6}
                        & \multirow{3}{*}{\emph{Multi-layer Perceptron}}  & \emph{No}            & 0.95      & 0.99   & 0.97     \\  
                               &                       & \emph{Yes}           & 0.57      & 0.22   & \textbf{0.31}     \\ 
                                &                      & \emph{Macro-average} & 0.76      & 0.60   & \textbf{0.64}     \\ \cline{2-6}
                        & \multirow{3}{*}{\emph{Random Forest}}  & \emph{No}            & 0.93      & 1.00   & 0.97     \\  
                                     &                 & \emph{Yes}           & \textbf{0.75}      & 0.03   & 0.06     \\ 
                                     &                 & \emph{Macro-average} & \textbf{0.84}      & 0.51   & 0.51     \\ \hline
\multirow{12}{*}{\emph{SBERT}} & \multirow{3}{*}{\emph{Logistic Regression}}   & \emph{No}            & 0.95      & 0.85   & 0.90     \\  
                                    &                  & \emph{Yes}           & 0.15      & \textbf{0.40}   & 0.20     \\  
                                    &                  & \emph{Macro-average} & 0.55      & \textbf{0.62}   & 0.55     \\ \cline{2-6}
                        & \multirow{3}{*}{\emph{SVM}}   & \emph{No}            & 0.95      & 0.98   & 0.96     \\  
                                 &                     & \emph{Yes}           & 0.50      & 0.25   & \textbf{0.33}     \\  
                                 &                     & \emph{Macro-average} & 0.72      & \textbf{0.62}   & \textbf{0.65}     \\ \cline{2-6}
                        & \multirow{3}{*}{\emph{Multi-layer Perceptron}} & \emph{No}            & 0.94      & 1.00   & 0.97     \\ 
                        &                              & \emph{Yes}           & 0.65      & 0.08   & 0.14     \\  
                        &                              & \emph{Macro-average} & 0.79      & 0.54   & 0.55     \\ \cline{2-6}
                        & \multirow{3}{*}{\emph{Random Forest}}         & \emph{No}            & 0.94      & 1.00   & 0.97     \\ 
                        &                              & \emph{Yes}           & \textbf{0.75}      & 0.08   & 0.14     \\  
                        &                              & \emph{Macro-average} & \textbf{0.84}      & 0.54   & 0.56     \\ \hline
\end{tabular}%
}
\label{tab:doccontainsframe2}
\end{table*}

\begin{table*}[!ht]
\tiny
\centering
\caption{Comparing \emph{Paragraph Vector} (\emph{PV-DM-NEG} and \emph{PV-DBOW-NEG}) based Classifiers for \emph{ParContainsFrame} in terms of \emph{Precision}, \emph{Recall} and \emph{F1-score} for each class (\emph{Yes} and \emph{No}) as well as macro-average across the classes}
\resizebox{\textwidth}{!}{%
\begin{tabular}{|c|c|c|c|c|c|}
\hline
\emph{Embedding} &   \emph{Classifier} & \emph{Class}         & \emph{Precision} & \emph{Recall} & \emph{F1-score} \\ \hline
\multirow{12}{*}{\emph{PV-DM-NEG}} & \multirow{3}{*}{\emph{Logistic Regression}} & \emph{No}    & 0.97      & 0.84   & 0.90 \\  
                                &                      & \emph{Yes}           & 0.25      & \textbf{0.71}   & 0.37     \\ 
                                &                      & \emph{Macro-average} & 0.61      & \textbf{0.77}   & 0.64     \\ \cline{2-6}
                                & \multirow{3}{*}{\emph{SVM}}       & \emph{No}            & 0.95      & 0.97   & 0.96     \\  
                                &                          & \emph{Yes}           & 0.40      & 0.29   & 0.34     \\  
                                &                          & \emph{Macro-average} & 0.67      & 0.63   & 0.65     \\ \cline{2-6}
                                & \multirow{3}{*}{\emph{Multi-layer Perceptron}} & \emph{No}    & 0.95      & 0.99   & 0.97     \\  
                                &                          & \emph{Yes}           & 0.61      & 0.29   & \textbf{0.39}     \\ 
                                &                          & \emph{Macro-average} & 0.78      & 0.64   & \textbf{0.68}     \\ \cline{2-6}
                                & \multirow{3}{*}{\emph{Random Forest}} & \emph{No}            & 0.93      & 1.00   & 0.96     \\  
                                &                          & \emph{Yes}           & \textbf{0.83}      & 0.01   & 0.01     \\ 
                                &                          & \emph{Macro-average} & \textbf{0.88}      & 0.50   & 0.49     \\ \hline
\multirow{12}{*}{\emph{PV-DBOW-NEG}} & \multirow{3}{*}{\emph{Logistic Regression}}    & \emph{No} &   0.98   &  0.84 &  0.91 \\  
                            &                           & \emph{Yes}           &   0.28    &  \textbf{0.80}  &   0.41   \\  
                            &                          & \emph{Macro Average} &   0.63    &  \textbf{0.82}  &  0.66    \\ \cline{2-6}
                            & \multirow{3}{*}{\emph{SVM}}   & \emph{No}            &   0.95    &  0.97  &  0.96    \\  
                            &                           & \emph{Yes}           &   0.48    &  0.40  &   0.43   \\  
                            &                          & \emph{Macro Average} &   0.72   &  0.68 &  0.70   \\ \cline{2-6}
                            & \multirow{3}{*}{\emph{Multi-layer Perceptron}}    & \emph{No}            &  0.95    &  0.99 &  0.97   \\  
                            &                          & \emph{Yes}           &  0.67    &  0.37 &   \textbf{0.47}  \\  
                            &                          & \emph{Macro-average} &   0.81   &  0.68 &  \textbf{0.72}   \\ \cline{2-6}
                             & \multirow{3}{*}{\emph{Random Forest}}        & \emph{No}            &  0.93    &  1.00 &  0.96   \\  
                             &                         & \emph{Yes}           &  \textbf{0.85}    &  0.01 &   0.03  \\  
                             &                         & \emph{Macro-average} &   \textbf{0.89}   &  0.51 &  0.49   \\ \hline
\end{tabular}%
}
\label{tab:parcontainsframe1}
\end{table*}

\begin{table*}[!ht]
\centering
\caption{Comparing \emph{Paragraph Vector} (\emph{PV-DM-HS} and \emph{PV-DBOW-HS}) and \emph{SBERT} based Classifiers for \emph{ParContainsFrame} in terms of \emph{Precision}, \emph{Recall} and \emph{F1-score} for each class (\emph{Yes} and \emph{No}) as well as macro-average across the classes}
\resizebox{\textwidth}{!}{%
\begin{tabular}{|c|c|c|c|c|c|}
\hline
\emph{Embedding} &   \emph{Classifier} & \emph{Class}         & \emph{Precision} & \emph{Recall} & \emph{F1-score} \\ \hline
\multirow{12}{*}{\emph{PV-DM-HS}} & \multirow{3}{*}{\emph{Logistic Regression}}  & \emph{No} &  0.97    &  0.81 &  0.88 \\  
                              &                        & \emph{Yes}           &  0.22    &  \textbf{0.73} &   0.34  \\  
                              &                        & \emph{Macro-average} &   0.60   &  \textbf{0.77} &  0.61   \\ \cline{2-6}
                              & \multirow{3}{*}{\emph{SVM}}                 & \emph{No}            & 0.96      & 0.94   & 0.95     \\  
                              &                       & \emph{Yes}           & 0.37      & 0.45   & 0.40     \\  
                              &                          & \emph{Macro-average} & 0.66      & 0.70   & 0.67     \\ \cline{2-6}
                              & \multirow{3}{*}{\emph{Multi-layer Perceptron}}  & \emph{No} & 0.95      & 0.98   & 0.97     \\  
                              &                          & \emph{Yes}           & 0.61      & 0.35   & \textbf{0.44}     \\  
                              &                          & \emph{Macro-average} & 0.78      & 0.67   & \textbf{0.70}     \\ \cline{2-6}
                              & \multirow{3}{*}{\emph{Random Forest}}        & \emph{No}            & 0.93      & 1.0   & 0.96     \\  
                              &                          & \emph{Yes}           & \textbf{1.0}      & 0.03   & 0.06     \\  
                              &                          & \emph{Macro-average} & \textbf{0.97}      & 0.52   & 0.51     \\ \hline
\multirow{12}{*}{\emph{PV-DBOW-HS}} & \multirow{3}{*}{\emph{Logistic Regression}}  & \emph{No}    & 0.98      & 0.82   & 0.89  \\ 
                                    &                  & \emph{Yes}           & 0.25      & \textbf{0.80}   & 0.38 \\ 
                                    &                  & \emph{Macro-average} & 0.62      & \textbf{0.81}   & 0.64 \\ \cline{2-6}
                                    & \multirow{3}{*}{\emph{SVM}}     & \emph{No}            &  0.96     &  0.96  &   0.96   \\  
                                    &                  & \emph{Yes}           &   0.49    &  0.54  &  \textbf{0.51}    \\  
                                    &                  & \emph{Macro-average} &   0.72    &  0.75  &  \textbf{0.73}    \\ \cline{2-6}
                                    & \multirow{3}{*}{\emph{Multi-layer Perceptron}} & \emph{No}    &   0.96    &  0.99  &   0.97   \\  
                                    &                  & \emph{Yes}           &    0.68   &  0.39  &   0.49   \\  
                                    &                  & \emph{Macro-average} &   0.82    & 0.69   &  \textbf{0.73}    \\ \cline{2-6}
                                    & \multirow{3}{*}{\emph{Random Forest}}  & \emph{No} &   0.93    & 1.00   &  0.96    \\  
                                    &                  & \emph{Yes}           &   \textbf{0.92}    &  0.04  &   0.08   \\  
                                    &                  & \emph{Macro-average} &    \textbf{0.92}   & 0.52   &  0.52    \\ \hline
\multirow{12}{*}{\emph{SBERT}} & \multirow{3}{*}{\emph{Logistic Regression}}  & \emph{No}    & 0.98      & 0.86   & 0.92  \\ 
                                    &                  & \emph{Yes}           & 0.31      & \textbf{0.79}   & 0.44 \\ 
                                    &                  & \emph{Macro-average} & 0.65      & \textbf{0.82}   & 0.68 \\ \cline{2-6}
                                    & \multirow{3}{*}{\emph{SVM}}     & \emph{No}            &  0.97     &  0.97  &   0.97   \\  
                                    &                  & \emph{Yes}           &   0.57    &  0.56  &  \textbf{0.57}    \\  
                                    &                  & \emph{Macro-average} &   0.77    &  0.77  &  \textbf{0.77}    \\ \cline{2-6}
                                    & \multirow{3}{*}{\emph{Multi-layer Perceptron}} & \emph{No}    &   0.95    &  0.99  &   0.97   \\  
                                    &                  & \emph{Yes}           &    0.72   &  0.35  &   0.46   \\  
                                    &                  & \emph{Macro-average} &   0.84    & 0.67   &  0.72    \\ \cline{2-6}
                                    & \multirow{3}{*}{\emph{Random Forest}}  & \emph{No} &   0.94    & 1.00   &  0.97    \\  
                                    &                  & \emph{Yes}           &   \textbf{0.82}    &  0.23  &   0.35   \\  
                                    &                  & \emph{Macro-average} &    \textbf{0.88}   & 0.61   &  0.66    \\ \hline
\end{tabular}%
}
\label{tab:parcontainsframe2}
\end{table*}

\begin{table*}[!ht]
\centering
\caption{Comparing Self-Attention based Classifiers (\emph{SelfAttention} and \emph{GuidedSelfAttention}) for \emph{ParContainsFrame} in terms of \emph{Precision}, \emph{Recall} and \emph{F1-score} for each class (\emph{Yes} and \emph{No}) as well as macro-average across the classes.}
\resizebox{\textwidth}{!}{%
\begin{tabular}{|c|c|c|c|c|}
\hline
\emph{Classifier} & \emph{Class} & \emph{Precision} & \emph{Recall} & \emph{F1-score} \\ \hline
\multirow{3}{*}{\emph{SelfAttention}}                 & \emph{No}            & 0.98      & 0.99   & 0.98     \\  
                                                      & \emph{Yes}           & \textbf{0.80}      & 0.67   & 0.72     \\  
                                                      & \emph{Macro-average} & \textbf{0.89}      & 0.83   & 0.85     \\ \hline
\multirow{3}{*}{\emph{GuidedSelfAttention}}           & \emph{No}            & 0.99      & 0.98   & 0.98     \\  
                                                      & \emph{Yes}           & 0.76      & \textbf{0.87}   & \textbf{0.81}     \\  
                                                      & \emph{Macro-average} & 0.88      & \textbf{0.93}   & \textbf{0.90}     \\ \hline
\end{tabular}%
}
\label{tab:parcontainsframedeep}
\end{table*}

\section{Acknowledgements}

This work was performed under contract with the Center for Security and Emerging Technology (\emph{CSET}) at Georgetown University, contract number CON-0011700. We would also like to thank \emph{SPi Global} for providing us AI Rhetorical Frame annotations at the document and paragraph level.

\bibliographystyle{ACM-Reference-Format}
\bibliography{rhetoricalFrame}

\end{document}